\documentclass{article}

\usepackage{arxiv}

\usepackage[utf8]{inputenc} 
\usepackage[T1]{fontenc}    
\usepackage{hyperref}       
\usepackage{url}            
\usepackage{booktabs}       
\usepackage{amsfonts}       
\usepackage{nicefrac}       
\usepackage{microtype}      
\usepackage{lipsum}		
\usepackage{graphicx}
\usepackage{natbib}
\usepackage{doi}
\usepackage{amsmath}
\usepackage{amssymb}
\usepackage{multirow}
\usepackage{subcaption}
\usepackage{float}

\title{Discovering Car-following Dynamics from Trajectory Data through Deep Learning}

\author{Ohay Angah\\
	Department of Civil and Environmental Engineering\\
	University of Washington\\
	Seattle, WA USA \\
	\And
	{James Enouen} \\
	Department of Computer Science\\
	University of Southern California\\
	Los Angeles, CA USA \\
        \And
        Xuegang (Jeff) Ban\\
	Department of Civil and Environmental Engineering\\
	University of Washington\\
	Seattle, WA USA \\
	\texttt{banx@uw.edu} \\
        \And
        {Yan Liu} \\
	Department of Computer Science\\
	University of Southern California\\
	Los Angeles, CA USA \\
}

\renewcommand{\shorttitle}{\textit{arXiv} Template}

\hypersetup{
pdftitle={A template for the arxiv style},
pdfsubject={q-bio.NC, q-bio.QM},
pdfauthor={David S.~Hippocampus, Elias D.~Striatum},
pdfkeywords={First keyword, Second keyword, More},
}

\begin{document}
\maketitle

\begin{abstract}
This study aims to discover the governing mathematical expressions of car-following dynamics from trajectory data directly using deep learning techniques. We propose an expression exploration framework based on deep symbolic regression (DSR) integrated with a variable intersection selection (VIS) method to find variable combinations that encourage interpretable and parsimonious mathematical expressions. In the exploration learning process, two penalty terms are added to improve the reward function: (i) a complexity penalty to regulate the complexity of the explored expressions to be parsimonious, and (ii) a variable interaction penalty to encourage the expression exploration to focus on variable combinations that can best describe the data. 
We show the performance of the proposed method to learn several car-following dynamics models and discuss its limitations and future research directions. 
\end{abstract}

\keywords{Car-following dynamics \and Deep symbolic regression \and Deep learning \and Variable interaction selection}

\section{Introduction}
\label{sec:intro}

There are two recent trends in transportation and the broader science/engineering fields, which make the headlines almost every day. The first one is the emergence of connected/automated vehicles (CAVs) that i) may introduce new, complex traffic dynamics and interactions in the current and future traffic streams, and ii) generate increasingly available and massive datasets from both vehicles and the infrastructure. The second trend is the rapid development and application of deep learning techniques that seem to revolutionize almost every aspect of technology, science, engineering, and the entire society. While there have been numerous studies and applications of deep learning in transportation, in the paper, we are interested in the question of whether deep learning can help discover traffic dynamics (car-following models in particular) from data directly with no or little human involvement. An affirmative answer to this question will not only help discover/develop traffic dynamics models in this era but also have important implications for other science/engineering fields where dynamical systems and their governing equations are widely used and studied.

Car-following depicts the driving behavior of how a vehicle (driver) follows and interacts with the vehicle in front of it. It is one of the basic traffic models in revealing traffic dynamics characteristics at the microscopic traffic flow level \cite{brackstone1999car}. Car-following studies can be traced back to the 1950s and 1960s when \citet{pipes1953operational, chandler1958traffic, kometani1958stability, gazis1959car, gazis1961nonlinear}, and \citet{helly1959simulation} initiated an era of modeling car-following and traffic dynamics. \citet{pipes1953operational} modeled car-following dynamics on the assumption that a following vehicle (\textit{the follower}) is required to maintain a certain distance from the preceding vehicle (\textit{the leader}). \citet{chandler1958traffic} further analyzed vehicle movements by a linear relationship between the acceleration of the follower and the velocity gap of the leader-follower pair. The model is given in Eq.~\eqref{equ:chandler} below. Here $a_{f}(t+\delta t)$, $v_l (t)$, and $v_f (t)$, respectively, represent the follower's acceleration at the next time step, and the velocity of the leader and the velocity of the follower at the current time step, with $c$ the so-called sensitivity term that may depend on the system states. 

\begin{equation}
    a_{f} (t+\delta t) = c[v_l (t) - v_f (t)].
    \label{equ:chandler}
\end{equation}

The model listed in Eq.\eqref{equ:chandler} is named the General Motors (GM) model and is widely regarded as the foundational model based on which many later car-following models were developed, e.g., by specifying various forms of the sensitivity term $c$. It states that the follower will accelerate or decelerate depending on the speed difference between the leader and the follower, while the action is with a time delay $\delta t$ representing the reaction time of the follower. The model is simple and intuitively makes sense, and was also validated by real-world data \cite{chakroborty1999evaluation}. It can thus be considered as the "governing equation" of the underlying dynamical system of car-following. The example of the above GM model also illustrates the unique feature of existing, physics-based car-following models: they are concise, explicit analytical models that have clear physical meanings and generalize well with different datasets/sites. 

The development of existing car-following models can be generally summarized as a scientific process \cite{treiber2002reconstructing, wen2007analysis, pourabdollah2017calibration, kurtc2020studying}, which includes key steps such as problem definition, background research, experimental design, data collection, data analysis, hypothesis, and testing/validation. This is often an involved, iterative process that is time and resource-consuming. The \textit{data analysis} and \textit{hypothesis} steps are arguably the most intellectually challenging and most creative in the entire process. This is because the \textit{data analysis} step requires innovative ways with intuition/knowledge to interpret the data, properly visualize, reveal salient patterns, and discover the right variables and parameters to build suitable car-following relations, as well as to apply knowledge to compare/contrast the patterns with existing models in the same or similar fields. This \textit{data analysis} step draws useful insights that serve as the input to the \textit{hypothesis} step. This later step then proposes a suitable dynamical model to best describe the discovered patterns and relations in the \textit{data analysis} step, while at the same time ensuring that the proposed model is tractable for analysis and computation. This \textit{data analysis}-\textit{hypothesis} process often involves many iterations for trial and error, with heavy human involvement.

With the advances in deep learning and emerging technologies in transportation, we wonder if the \textit{data analysis}-\textit{hypothesis} steps in the existing scientific process can be replaced by a deep learning approach with minimum human interaction, for discovering the governing equations of car-following dynamics. The motivation for this is multifaceted. First, while predictive models \cite{chen2021discovering, kamienny2022end, shi2021physics}, including those predicting the car-following behavior \cite{mo2021physics}, have been increasingly developed in transportation based on deep learning techniques and outperformed traditional physics-based models, they are often a black box that lacks interpretations. We believe that concise models with clear physical interpretations, as traditionally done in transportation and other fields, are still crucial as "human knowledge" and in understanding and modeling new phenomena with emerging technologies and systems. This leads to the second motivation, i.e., important practical needs to develop such learning-based methods. Existing car-following models were developed mainly for human-driven vehicles (HDVs). With the emergence and rapid deployment of CAVs, it is imperative to understand the interactions, including car-following, among CAVs and between CAVs and HDVs \cite{zhong2020influence, xiao2018unravelling}. Distinct to CAVs, CAV and HDV data may be collected by CAV themselves or from the infrastructure \cite{gloudemans202324}, which may make the classical scientific process cumbersome to discover the governing equations for new car-following and other traffic dynamics. Deep learning models may be a promising alternative method. Third, it is an intriguing research topic to develop a fully data-driven approach integrated with deep learning, with no or little human involvement, to directly produce a car-following model that shares the same features with the existing models: concise, interpretable, and generalizable. In the meantime, a successful development of such learning-based methods can provide useful insights into the interpretability of deep learning techniques, an open question that is drawing much attention nowadays.

Developing the aforementioned learning techniques for car-following from data can be viewed as a data-driven discovery problem \cite{rudy2017data, brunton2016discovering, antoniou2011synthesis}. Early research has been primarily investigated on how to draw physical laws from ideal scientific experiments using classical methods such as regression models. In the early 2000s, \citet{bongard2007automated} and \citet{schmidt2009distilling} proposed an approach named "symbolic regression (SR)" for identifying the underlying governing equation of a dynamical system directly from data. The purpose of the SR approach is to determine a mathematical expression $f: \mathbb{R}^d \rightarrow \mathbb{R}$, $X_i \in \mathbb{R}^d$, $y_i\in \mathbb{R}$, and a dataset $(X, y),$ such that $f$ is a short mathematical expression and $y \approx f(X)$. The SR model was solved by Genetic Programming (GP) \citep{koza1994genetic} based approaches, which were later widely extended \citep{bongard2007automated, schmidt2009distilling, back2018evolutionary, la2016epsilon, la2018learning, virgolin2019linear, virgolin2021improving, kommenda2020parameter, de2021interaction, petersen2019deep}. We refer the readers to Section \ref{sec:literature} for more detailed reviews.

Despite the above progress of learning-based discovery methods, discovering the underlying governing expressions from real-world traffic data that often involves noise and uncertainties can be challenging. Further, integrating with GP to solve SR problems may struggle with the downside of GP, such as sub-optimal and complex solutions. This study applies deep symbolic regression (DSR) \cite{petersen2019deep} and GP to uncover the underlying car-following mathematical models from trajectory data. In particular, we make two improvements to the DSR-GP framework. First, DSR-GP is given some prior knowledge before learning the dynamics. The prior knowledge is a set of feature (variable) combinations extracted from the trajectory data through the variable intersection selection (VIS) algorithm proposed by \citet{enouen2022sparse}. Through this prior knowledge, the framework is guided to mathematical expressions that have reasonable physical meanings. Second, we propose a new reward function to encourage exploring low-complexity mathematical expressions. These two improvements can lead to expressions that are parsimonious and interpretable. We call the proposed learning framework the VIS-enhanced DSR.

In the next section, we briefly review the DSR and VIS methods. The VIS-enhanced DSR framework is presented in Section~\ref{sec:method}. In this framework, we first apply VIS to discover variable combinations from a trajectory dataset and then use DSR to learn the dynamics based on the discovered variable combinations. The expression reconstruction performance of the proposed framework is compared with the performance using only DSR and DSR with GP. Further, we investigate the impact of data noise and the performance of the method on different types of car-following dynamics. The results and comparisons are included in Section~\ref{sec:results}. We discuss the limitation of the proposed method in Section~\ref{sec:discussion}, with conclusions and future research directions presented in Section~\ref{sec:conclusion}.

The main contributions of this study are summarized as follows.
\begin{itemize}
  \item Deep learning models are proposed for identifying car-following dynamics from trajectory data with minimum human involvement.
  \item VIS-enhanced DSR framework is developed to find mathematical expressions that tend to be parsimonious and interpretable.
  \item Reward function in DSR is improved by new penalty terms to regulate the complexity of the explored expressions and to search expressions with desirable feature combinations. 
\end{itemize}

\section{Related Works}
\label{sec:literature}

In this section, we briefly review car-following models and the data-driven models in solving SR problems, including the GP methods and the DSR framework \cite{petersen2019deep}. We also briefly review the VIS approach \cite{enouen2022sparse}. 

\subsection{Car-following Models}

Most of the pioneering car-following models \cite{kometani1958stability, pipes1966car, herman1959traffic, chandler1958traffic, edie1961car, helly1959simulation, gazis1959car, gazis1961nonlinear} between fifties to seventies modeled car-following dynamics as an acceleration function with the differences of positions and speeds between the leader and the follower. Later on, car-following models were extended to consider multiple scenarios including the follower's emergency braking, following the leader, approaching the leader, free moving, and accelerating from rest, where notable works include  \cite{wiedemann1992microscopic, michaels1963perceptual, burnham1976heuristic, lee1976theory, kumamoto1995rule, krauss1998microscopic, treiber2000congested}. In particular, the Krauss model \cite{krauss1998microscopic} considers that the follower selects its action from its desired speed, its comfortable acceleration, or the speed that is able to maintain a safe distance from the leader. The Intelligent Driver Model (IDM) \citep{treiber2000congested} further considers a "soft braking strategy" when the follower approaches a slow-moving leader. It has been shown that the Krauss model performs better over IDM when modeling unsteady traffic conditions, whereas IDM performs better under steady traffic conditions \cite{kanagaraj2013evaluation, naveen2013microscopic}. In the numerical experiments of this paper, we use a traffic simulation model that integrates the Krauss model for car following. The model is defined by Eq.~\eqref{equ:modified-krauss}, where $t_{\text{react}}, v_f(t), v_l(t), s_f(t), \text{ and } s_{\text{des}}$ respectively represent follower's reaction time, the velocities of the follower and the leader at the current time step, the spacing at the current time step, and the desired spacing ($s_{\text{des}}$) that is defined by measuring the distance that the leader is estimated to drive forward in the next time interval, i.e., $s_{\text{des}}=v_l(t)\delta t$. The equation describes that given the current velocities of the leader and the follower, and the spacing between the pair, the follower selects its velocity from the choices of its current velocity plus its maximum acceleration $v_f (t)+a_{\text{max}}\delta t$, safe velocity $v_s$, and the maximum velocity that this vehicle is capable of driving at, i.e., $v_{\text{max}}$ that is a known parameter. Other parameters that are pre-defined and assumed fixed in this model include the maximum acceleration $a_{\text{max}}$, comfortable deceleration $b$, driver's reaction time $t_{\text{react}}$, driving uncertainty $\varepsilon$, and time step $\delta t$. 

\begin{equation}
\begin{aligned}
    v_f\ (t+\delta t)&=\max(0,\ v_d-\varepsilon a_{\text{max}}) \\
    v_d&=\min(v_f (t)+a_{\text{max}}\delta t, v_s, v_{\text{max}}) \\
    v_s&=v_l (t)+\frac{s_f(t)-s_{\text{des}}(t)}{ \frac{v_f (t)+v_l (t)}{2b}+t_{\text{react}}}.
\end{aligned}
\label{equ:modified-krauss}
\end{equation}

As this paper concerns about discovering existing car-following models using deep learning methods, we do not aim to conduct a comprehensive review of car-following models; interested readers may refer to \cite{brackstone1999car, treiber2013traffic} for more recent reviews.

\subsection{Data-driven Discovery Methods}

GP \cite{koza1994genetic} was used to solve SR problems \cite{bongard2007automated, schmidt2009distilling}, which was shown to be effective by leveraging the evolutionary features of GP to improve fitting functions \citep{koza1994genetic, schmidt2009distilling, back2018evolutionary}. Many later approaches for solving SR problems were based on the concept of GP, including the epsilon-lexicase selection (EPLEX) \cite{la2016epsilon}, the feature engineering automation tool (FEAT) \cite{la2018learning}, semantic backpropagation GP (SBP-GP) \cite{virgolin2019linear}, the GP-based gene-pool optimal mixing evolutionary algorithm (GP-GOMEA) \cite{virgolin2021improving}, Operon \cite{kommenda2020parameter}, and the interaction-transformation evolutionary algorithm (ITEA) \cite{de2021interaction}. Despite the powerful solving capabilities from the nature of evolution, GP methods suffer from generating appropriate expressions that are parsimonious and general \citep{o2010open}, and fitting may take a long time due to the large size of search space.

\citet{brunton2016discovering} developed the sparse identification of nonlinear dynamics (SINDy) framework that describes the expression discovery problem by a sparse regression problem. The framework is based on an underlying assumption that the dynamics of a physical system can be represented by a linear combination of a limited number of mathematical terms. The SINDy approach was integrated with autoencoders by \citet{champion2019data} and further with Bayesian approaches by \citet{gao2022bayesian}. However, SINDy users need to provide all the possible feature combinations for the framework to discover important terms. \citet{jin2019bayesian} incorporated Bayesian approaches in solving SR problems. In the deep learning domain, the AIFeynman approach \citep{udrescu2020ai} is also worth mentioning. It is inspired by the divide-and-conquer method powered by neural networks. \citet{cranmer2020discovering} and \citet{kamienny2022end} applied graph neural networks (GNNs) and transformers to solve SR problems, respectively. Similar to the works by \citet{champion2019data} and \citet{gao2022bayesian}, which discovered the governing behaviors from the intrinsic coordinates, \citet{chen2021discovering} proposed a two-stage autoencoder framework to estimate the intrinsic variables and use the variables to predict the behaviors at the next time step. The intrinsic patterns are however hidden in the autoencoders and thus are not interpretable. 

DSR \cite{petersen2019deep} is a recent framework that uses a recurrent neural network (RNN) with the proposed risk-seeking policy gradient approach for solving SR problems. DSR leverages the conditional probability feature in RNNs to increase the parsimony and accelerate the exploration efficiency. It was developed on the basis of the fact that a mathematical expression can be represented by an expression tree \citep{meurer2017sympy} composed of mathematical tokens, where the tokens are sampled by RNNs from a pre-defined token pool. The sampling is restricted by fixed constraints to restrict token combinations and lengths of expressions, i.e., expression complexities. Given a range of expression complexity as an exploration constraint, the RNN will stop sampling if the expression has enough complexity and the expression tree is complete. This mechanism may result in expressions with a high level of complexity that lose parsimony and interpretability, which is also the main issue of GP. Data noise is another challenge: results showed that 10\% of data noise affected over 10\% of expression reconstruction, although it outperformed benchmark solvers, including GP, Eureqa \cite{schmidt2009distilling}, etc. Expression searching in the DSR framework relies on training RNNs or long-short-term-memory (LSTM) neural networks, a special structure of RNNs, whose training can sometimes be computationally expensive. Combining GP with DSR has also been proposed \cite{larma2021improving}, for which GP is an option to assist the search. 

\subsection{Variable Interaction Selection (VIS)}
\label{subsec:VIS}
To encourage the exploration to search for expressions that have reasonable lengths and physical meanings, we add a VIS step prior to DSR to provide knowledge to the exploration. This VIS step produces potential variable combinations extracted from data. The VIS algorithm was introduced in \citet{enouen2022sparse} in their proposed sparse interaction additive networks (SIANs). SIANs leverages the generalized additive function, which speculates that a dependent variable is linearly related to an addition of a set of shape functions. A shape function can be any specific function, e.g., a polynomial function of a single variable or variable combinations that have a strong interaction, where the interaction strength is quantified by Archipelago \cite{tsang2020does}. 
The algorithm first fits a deep neural network to the data, where the deep neural network acts as the reference function. The algorithm detects all the possible variable combination candidates and the selection is based on several key criteria including a pre-defined threshold of the interaction strength. That is, a variable combination is determined as a strong interaction if its interaction strength is greater than the threshold. We next present the VIS-enhanced DSR method to discover car-following models from data.

\section{VIS-Enhanced DSR Method}
\label{sec:method}

\begin{figure}[H]
    \centering
    \begin{subfigure}{.5\textwidth}
      \centering
      \includegraphics[width=\linewidth]{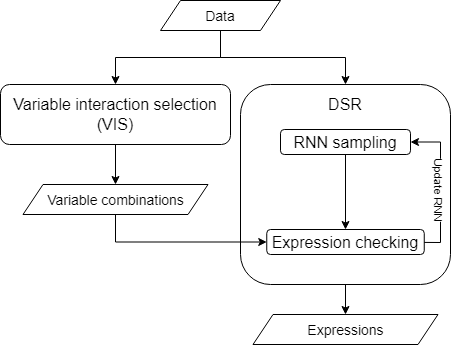}
      \caption{Expression Exploration Flowchart}
      \label{fig:overall-flowchart}
    \end{subfigure}%
    \begin{subfigure}{.5\textwidth}
      \centering
      \includegraphics[width=0.8\linewidth]{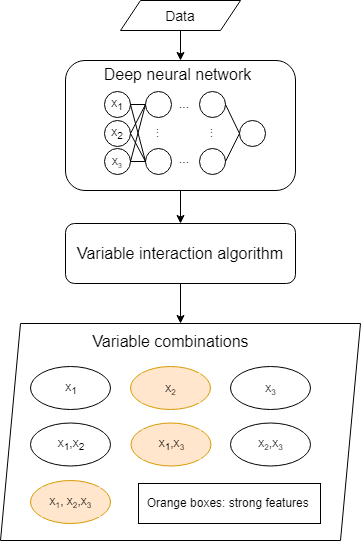}
      \caption{Variable Interaction Pipeline \cite{enouen2022sparse}}
      \label{fig:VIS-pipeline}
    \end{subfigure}
    \caption{Expression Exploration Flowchart And An Example}
    \label{fig:pipeline}
\end{figure}

Fig.~\ref{fig:overall-flowchart} depicts the flowchart of our proposed VIS-enhanced DSR method. On the left of the flowchart, we use a VIS process to detect potential variable combinations from the trajectory data to increase the interpretability of the explored expressions. On the right of the flowchart, we use an expression checker to detect whether the expression determined by the RNN sampling has reasonable physical meanings based on the recommended variable combinations. The RNN training is governed by our proposed reward function that encourages parsimonious expressions for better interpretation. Fig.~\ref{fig:VIS-pipeline} illustrates a pipeline example of the VIS process. Given a set of trajectory data, the process trains a deep neural network to match the data and then produces a set of variable combinations that can capture the salient patterns in the data. The identified variable combinations can guide DSR to search for mathematical expressions that contain such variables. To introduce the framework and the reward function in detail, we first define the problem and notations as follows. 

Define the same mathematical expression $f(X)$ and dataset $(X, y)$ introduced in Section~\ref{sec:intro}, i.e., 

\begin{equation}
    y=f(X)+\epsilon,
\end{equation}
where $\epsilon$ represents the approximation error term. We aim to discover the exact form of $f(X)$ from the dataset. Given a set of mathematical tokens $\mathcal{T}$, which can include mathematical operations ($+$, $-$, $sin$, constant placeholders, etc.) and variables, the DSR framework samples each mathematical token $\tau_i$ from $\mathcal{T}$ according to the probabilities determined by the RNN. The token in $\mathcal{T}$ with the highest probability will be sampled and stored in the latest node spot. Each mathematical term is sampled without replacement. We denote the resulting expression by $f_s(\tau)$, where $\tau = \{ \tau_i \}_{i=1} ^{n_s}, \ \forall \tau_i \in \mathcal{T}$, and $n_s$ represents the number of sampling occasions, i.e., expression complexity. We apply the DSR approach to find $f_s(\tau)$ through sampling tokens from the given $\mathcal{T}$ such that $\epsilon$ is small. As an example, $f_s(\tau)$ can be $\alpha x+sinx$ if the sample token set is \{$+$, $\times$, $\alpha$, $x$, sin, $x$\}, where $x$ is the only variable and $\alpha$ is a constant.

As shown in Fig.~\ref{fig:overall-flowchart}, we first feed the collected trajectory dataset, $(X, y)$, into both the VIS and DSR blocks. VIS starts to find the variable interactions from the trajectory data, as shown in Fig.~\ref{fig:VIS-pipeline}, by first fitting a deep neural network to the data, calculating the interaction strength for variables or variable combinations, and determining whether a variable or combination is significant to the data relationship. In the meantime, DSR starts to explore potential expressions that best describe the relationship between $X$ and $y$. The potential expressions are then fed into the expression-checking process, which will check each expression candidate term by term for reward evaluation. The checking process will also assess whether the expression has recommended variable combinations by VIS and tag those without any recommended combinations (so that a penalty may be imposed as shown in Section \ref{section:dsrCarfollowing}). The DSR process will then continue evaluating the reward of the retained expression candidates, training the RNN, and exploring potential expressions.

\subsection{Variable Interaction Selection (VIS)}

In the VIS process, we consider an extended generalized additive model \cite{enouen2022sparse}, in which the dependent variable linearly depends on some functions consisting of single variables and/or groups of variable combinations, as defined in Eq.\eqref{equ:y}. In the equation, $f_\varnothing$ is a constant term. Each $h_i(X_i)$ function is a shape function that represents a specific function, e.g., a polynomial function. $X_i$ is an element of the variable set, i.e., $X_i \in \mathcal{I} \subseteq \{X_i\}_{i=1}^{d}$. The last term considers the variable combinations, where $Z_j$ (e.g., $Z_j=(X_1, X_2)$) represents an element in the set of variable combinations. The variable combination set is defined as $\mathcal{J}$, $Z_j \in \mathcal{J}$.

\begin{equation}
    y = f_\varnothing + \sum_{X_i \in \mathcal{I}} h_i(X_i) + \sum_{Z_j \in \mathcal{J}} h_j(Z_j).
    \label{equ:y}
\end{equation}

In VIS, a deep neural network is first trained to fit the data. We use $f^{\text{dnn}}(X)$ to denote the deep neural network. The interaction strengths of variables (in set $\mathcal{I}$) or variable combinations (in set $\mathcal{J}$), denoted as $\psi(\mathcal{I})$ and $\psi(\mathcal{J})$ respectively, can be measured by the Archipelago \cite{tsang2020does}, which uses the approximation of the Hessian of $f^{\text{dnn}}(X)$ as shown in Eq.~\eqref{equ:formal-psi}. The VIS algorithm proposed in \cite{enouen2022sparse} is then used to detect strong variable interactions performed in the data based on $f^{\text{dnn}}(X)$ and Eq.~\eqref{equ:formal-psi}. 

\begin{equation}
\begin{aligned}
    & \psi(\mathcal{I}) := \mathop{\mathbb{E}}_X \big[ \frac{\partial^{|\mathcal{I}|} f^{\text{dnn}}(X)}{\partial X_{i_1} \partial X_{i_2} \cdots \partial X_{i_{|\mathcal{I}|}} } \big]^2 > 0. \\
    & \psi(\mathcal{J}) := \mathop{\mathbb{E}}_X \big[ \frac{\partial^{|\mathcal{J}|} f^{\text{dnn}}(X)}{\partial Z_{j_1} \partial Z_{j_2} \cdots \partial Z_{j_{|\mathcal{J}|}} } \big]^2 > 0.
    \label{equ:formal-psi}
\end{aligned}
\end{equation}

\subsection{Improved DSR for Discovering Car-following Dynamics}\label{section:dsrCarfollowing}

In the DSR process, we use an LSTM neural network to sample tokens. The LSTM is trained through the risk-seeking policy gradient algorithm proposed by \citet{petersen2019deep}, which updates the LSTM weights through the top-performing cases (instead of the average-performing cases as normally done). We redesign the original reward function in \citet{petersen2019deep} to encourage the LSTM to (i) explore expressions consisting of the recommended variable combinations generated by VIS, and (ii) explore expressions with reasonable expression complexity. The new reward function is shown in Eq.~\eqref{equ:reward}, where $R_c$ represents the combined reward function, $L_e$ represents an error-based loss function, $p$ is the complexity of an expression $f_s$, and $\beta$ is a penalty parameter that penalizes expressions that do not contain the recommended variable combinations from VIS. The penalty only steps in for the first \textit{EPOCH} epochs to initially guide the exploring direction and prevent the exploration from being time-consuming, as checking each term of an expression needs extra time.

\begin{equation}
    \begin{aligned}
    R&= 
    \begin{cases}
        R_c(1-\beta),& \text{if } \text{no potential interactions and epoch} \leq \text{EPOCH}\\
        R_c,              & \text{otherwise},
    \end{cases}, \\
    R_c&= \frac{2}{(1+L_{e})+(1+\text{norm}(p(f_s)))}, \\
    \end{aligned}
    \label{equ:reward}
\end{equation}


As shown in Eq.~\eqref{equ:reward}, the combined reward function $R_c$ considers the error-based loss and the complexity of the expression candidate. $R_c$ is squashed to a range between 0 to 1. $R_c \rightarrow 1.0$ when the error loss and the loss of complexity are both close to zero, i.e., $L_e \rightarrow 0$ and $\text{norm} (p(f_s)) \rightarrow 0$. $R_c$ goes down when the predicted error and/or the complexity of an expression increase. To encourage the LSTM to explore expressions with reasonable complexities, we design the loss of expression complexity as a normalization function. The normalization relies on human knowledge to provide a reasonable range, as defined in Eq.~\eqref{equ:normalization}, where $p_{\min}$ and $p_{\max}$ represent the minimum expression complexity and maximum expression complexity, respectively, which are pre-defined based on human knowledge.

\begin{equation}
    \begin{aligned}
    \text{norm}(p(f_s))&= 
    \begin{cases}
        \frac{p(f_s) - p_{\min}}{p_{\max} - p_{\min}},& \text{if } p_{\min} \leq p(f_s)\\
        \frac{p_{\max} - p(f_s)}{p_{\max} - p_{\min}},              & \text{otherwise}.
    \end{cases}
    \end{aligned}
    \label{equ:normalization}
\end{equation}

The equation's expression complexity is scaled into a base range of 0 to 1. When the complexity is greater than or equal to the minimum $p_{\min}$, it is scaled by the top conversion of the equation. To prevent the same conversion from scaling the complexity to a negative number when the complexity is smaller than the minimum, we use the bottom conversion of the equation to scale the complexity. The bottom conversion comes from the idea of treating the smaller complexity (smaller than the desired range) as the greater complexity (greater than the desired range) to prevent producing negative losses. That is, if $p(f_s) < p_{\min}$, then $\text{norm}(p(f_s)) = \frac{\big( p_{\max} + (p_{\min} - p) \big) - p_{\min}}{p_{\max} - p_{\min}}$, and rewriting $\frac{\big( p_{\max} + (p_{\min} - p) \big) - p_{\min}}{p_{\max} - p_{\min}} = \frac{\big( p_{\max} - p \big)}{p_{\max} - p_{\min}}$ obtains the bottom conversion of Eq.~\eqref{equ:normalization}.

$L_e$ here captures the "accuracy" of the expression in describing the data. It can be any predicted loss function, such as the root-mean-squared error (RMSE), mean-squared-error (MSE), normalized MSE (NMSE), NRMSE, log MSE, etc. This study uses NRMSE as $L_e$. Given the expression formed by sampled tokens $f_s(\tau)$ and dataset $(X, y)$, $L_e$ is defined as Eq.~\eqref{eq:error-based-loss}, where $\sigma_y$ is the standard deviation of $y$.

\begin{equation}
    L_e = \frac{\sqrt{\frac{1}{n} \Sigma^n_{i=1} (y_i - f_s(\tau, X_i))^2}}{\sigma_y}.
    \label{eq:error-based-loss}
\end{equation}

Before concluding this section, we summarize the human involvements/knowledge that are required by the proposed VIS-enhanced DRS method. A set of tokens is first required, including variables ($X$) and mathematical operations. The variables can be relatively easily obtained as the key elements in the data: e.g., for trajectory data, this can be speeds, accelerations, locations, etc. Operations are a bit trickier and require certain level of human knowledge, e.g., the operations used in existing car-following models if new car-following models are to be developed. The good news is that we do not need to provide the exact set of operations, e.g., multiplication and subtraction for the car-following model in Eq. \eqref{equ:chandler}. Rather a \textit{superset} of the needed operations is sufficient, e.g., the set of operations that contain $\times, \div, +, -$ for Eq. \eqref{equ:chandler}. Other human inputs include the min and max complexities in Eq \eqref{equ:normalization} and the $\beta$ parameter in Eq \eqref{equ:reward}. These may be treated as hyberparameters for machine leanring, which may be pre-defined based on knowledge or learned by trial and error. Compared with traditional methods of deriving a car-following model from observed data, we can clearly see that the human involvement in the proposed method is marginal since the data analysis and hypothesis steps are largely automated via the VIS and DSR processes.

\section{Numerical Results}
\label{sec:results}

We now use the proposed VIS-enhanced DSR framework to help discover the potential governing mathematical expressions from car-following trajectory datasets. We first present how the datasets are prepared. 

\subsection{Dataset and Problem Statement}
\label{subsec:dataset}

The trajectory data is collected through numerical simulation with car-following pairs using the Simulation of Urban MObility (SUMO). To enhance the variety of the car-following behaviors, we control the leading vehicle by randomly selecting the velocity between 0 to 30 m/s, whereas the decisions of the following vehicle are described by the Krauss model \cite{krauss1998microscopic, krajzewicz2002sumo}, as defined in Eq.~\eqref{equ:modified-krauss}, with predefined vehicle parameters: $a_{\text{max}}=2.6 \text{ m/s}^2$, $v_{\text{max}}=55.55 \text{ m/s}$, $b=4.5 \text{ m/s}^2$, and 5-m vehicle length. The follower is first assumed to have zero driving uncertainty, i.e., $\varepsilon = 0.0$, and their reaction time is a second, i.e., $t_{\text{react}}=1.0$ second. Later, we also impose data noises and test how the proposed method can deal with such noises. The trajectory dataset has in total $\sim$3.6k car-following samples (pairs). Given the parameters, the Krauss model can be simplified as Eq.~\ref{equ:modified-krauss-simplified}. The function will be the target expression in the following experiments. From the simulation, we collect the velocities of the follower and the leader at the current time step, the spacing at the current time step, the following velocity at the next time step, and the difference between current spacing and the desired spacing, i.e., $, v_f(t)$, $v_l(t)$, $s_f(t)$, $v_f(t+\delta t)$, and $ds(t)=s_f(t)-s_{\text{des}}(t)$, where $s_{\text{des}}(t)=v_l(t)\delta t$.

The data is organized into independent variables $X=\{v_f(t), v_l(t), s_f(t), ds(t)\}$, whereas the dependent variable is the velocity of the follower at the next time step, i.e., $y=v_f(t+\delta t)$. The problem is to develop a mathematical expression that describes the relationship between the collected features and the next velocity of the follower, i.e., $v_f(t+\delta t) = f(v_f(t), v_l(t), s_f(t), ds(t))$.

\begin{equation}
\begin{aligned}
    v_f\ (t+\delta t)&=\min(v_f (t)+2.6, v_s) \\
    v_s&=v_l (t)+\frac{9.0 \big( s_f(t)-s_{\text{des}} (t) \big)}{ {v_f (t)+v_l (t)}+9.0}.
\end{aligned}
\label{equ:modified-krauss-simplified}
\end{equation}

Note that certain simplifications are performed to derive \eqref{equ:modified-krauss-simplified}, including only vehicles from a single class (cars), setting the reaction time to be 1s, as well as fixing the maximum acceleration and speed. In Section \ref{sec:discussion}, we discuss these issues and possible ways to relax them.

\subsection{Variable Interactions}
\label{subsec:vis-result}
\begin{figure}[H]
    \centering
    \includegraphics[width=\linewidth]{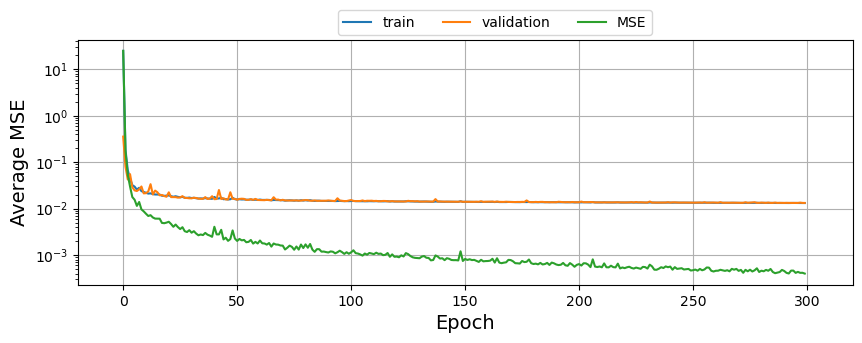}
    \caption{DNN Training Loss}
    \label{fig:dnn-loss}
\end{figure}

The interaction selection process starts by training a 3-layer (in-32-64-32-out) neural network. The trajectory dataset is split into an 80\% training set and a 20\% validation set. After training, the mean squared error of the deep neural network estimation drops to 7.5E-04, 7.1E-04, and 4.6E-04 after 50, 100, and 300 epochs of training, as shown in Fig.~\ref{fig:dnn-loss}. The training takes 5.8, 16.9, and 60.6 seconds at 50, 100, and 300 epochs on an Intel Core i7-12700 CPU. 

\begin{figure}[H]
    \centering
    \includegraphics[width=0.65\linewidth]{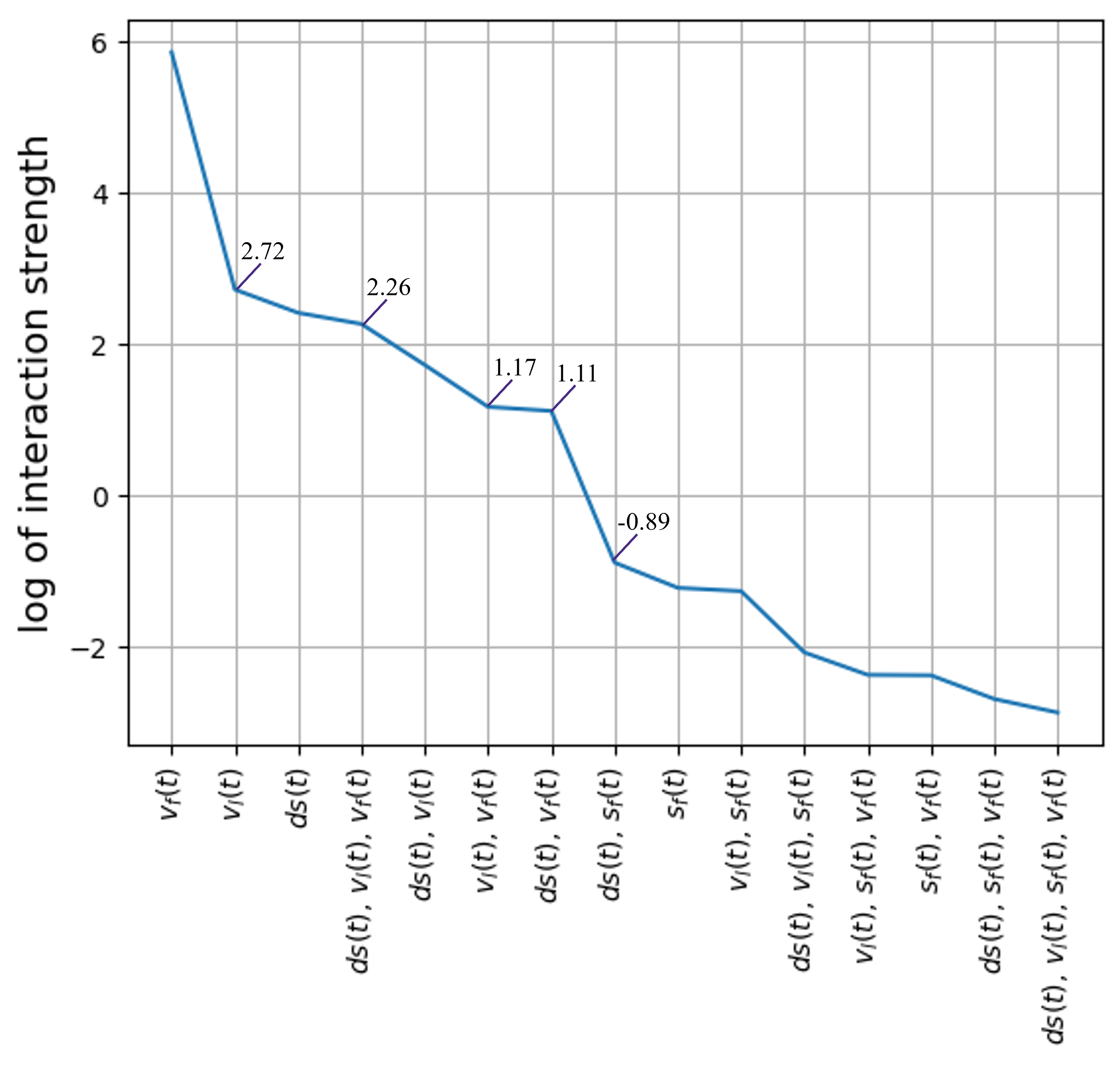}
    \caption{Interaction Strength Elbow Plot}
    \label{fig:interaction-strength}
\end{figure}

Given variables $v_f(t)$, $v_l(t)$, $s_f(t)$, and $ds(t)$, the potential variable combinations along with their interaction strengths after 300 epochs of training are listed in Table~\ref{tab:interactions}, where variable combinations are listed in ascending order by the interaction strength presented in Fig.~\ref{fig:interaction-strength}. In this figure, a few elbow points can be observed, including strengths of $v_l(t)$, \big($ds(t)$, $v_l(t)$, $v_f(t)$\big), \big($v_l(t)$, $v_f(t)$\big), \big($ds(t)$, $v_f(t)$\big), and \big($ds(t)$, $s_f(t)$\big). Based on these elbow points, five sets of variables or variable combinations are recommended as follows. 

\begin{enumerate}
    \item[Set 0] \{$v_f(t)$, $v_l(t)$\},
    \item[Set 1] \{$ds(t)$, \big($ds(t)$, $v_l(t)$, $v_f(t)$\big)\},
    \item [Set 2] \{\big($ds(t)$, $v_l(t)$\big), \big($v_l(t)$, $v_f(t)$\big)\},
    \item [Set 3] \{\big($ds(t)$, $v_f(t)$\big)\},
    \item [Set 4] \{\big($ds(t)$, $s_f(t)$\big)\}.
\end{enumerate}

The two variables in Set 0 are unlikely to be the only features (variables) describing the car-following behavior since the spacing difference may also be an important feature. We therefore combine Set 0-1 as the first possible recommended set of variable combinations, denoted as \textit{Scenario 1}. The rest sets are respectively coded up as \textit{Scenario 2} to \textit{Scenario 4}. Table~\ref{tab:interactions} presents all variable combinations, their interaction strengths, and the scenarios. For instance, Scenario \#1 uses Set 0 and Set 1, and Scenario \#2 uses Set 0 - Set 2. 

\begin{table}[H]
\scriptsize 
\centering    
\caption{Variable Interaction Strengths}
\begin{tabular}{ccccc}
Interaction                                  & Interaction Strength (IS) & $\log$(IS) & Scenario           & \multicolumn{1}{c}{Set} \\ \cline{1-5}
$v_f(t)$                                     & 347.07                    & 5.85       & \multirow{4}{*}{\#1} & \multirow{4}{*}{0 \& 1} \\
$v_l(t)$                                     & 15.12                     & 2.72       &                    &                         \\
$ds(t)$                                      & 11.12                     & 2.41       &                    &                         \\
$ds(t)$, $v_l(t)$, $v_f(t)$                  & 9.60                      & 2.26       &                    &                         \\ \cline{1-5}
$ds(t)$, $v_l(t)$                            & 5.59                      & 1.72       & \multirow{2}{*}{\#2} & \multirow{2}{*}{0 - 2} \\
$v_l(t)$, $v_f(t)$                           & 3.21                      & 1.17       &                    &                         \\ \cline{1-5}
$ds(t)$, $v_f(t)$                            & 3.04                      & 1.11       & \#3                  & 0 - 3                  \\ \cline{1-5}
$ds(t)$, $s_f(t)$                            & 0.41                      & -0.89      & \#4                  & 0 - 4                       \\ \hline
$s_f(t)$                                     & 0.29                      & -1.22      & \multicolumn{2}{c}{-}                        \\
$v_l(t)$, $s_f(t)$                           & 0.28                      & -1.27      & \multicolumn{2}{c}{-}                        \\
$ds(t)$, $v_l(t)$, $s_f(t)$                  & 0.13                      & -2.08      & \multicolumn{2}{c}{-}                        \\
$v_l(t)$, $s_f(t)$, $v_f(t)$                 & 0.09                      & -2.37      & \multicolumn{2}{c}{-}                        \\
$s_f(t)$, $v_f(t)$                           & 0.09                      & -2.38      & \multicolumn{2}{c}{-}                        \\
$ds(t)$, $s_f(t)$, $v_f(t)$                  & 0.07                      & -2.67      & \multicolumn{2}{c}{-}                        \\
$ds(t)$, $v_l(t)$, $s_f(t)$, $v_f(t)$        & 0.06                      & -2.87      & \multicolumn{2}{c}{-}                        \\ \hline
\end{tabular}
\label{tab:interactions}
\end{table}

\subsection{Expression Exploration}
Based on the literature and analysis in the past \citep{pipes1953operational, chandler1958traffic,gazis1959car,gazis1961nonlinear,helly1959simulation,pipes1966car,krauss1998microscopic,treiber2013car}, car-following behaviors are related to the velocity and gap differences with $+$, $-$, $\times$, $\div$, $\min$ and parameters $a_{\text{max}}$, and $b$. We thus provide $+$, $-$, $\times$, $\div$, $\min$, $v_f(t)$, $v_l(t)$, $ds(t)$, $s_f(t)$, $a_{\text{max}}$, and $b$ as the set of tokens to the DSR process. If we allow exploration to focus solely on fitting accuracy without constraints and complexity regulations, we may end up with a complex expression that lacks physical meaning, e.g., the expression below, which may have a high level of accuracy but also a high level of expression complexity. Such expression is certainly not desirable.

\begin{equation*}
\footnotesize
\begin{split}    
        (s_f(t)\left(-v_l(t)/\left(\left(v_f(t)\ -\ a_{\text{max}}/\left(v_f(t) s_f(t)\ \min\{-2b,\ v_l(t)\}\right)\right) \min\{v_f(t),\ v_l(t)\}\right)\ -\ 2b\right)\ + \\ \ s_f(t)\ -\min\{a_{\text{max}},\ s_f(t)\}\ -\min\{v_f(t),\ a_{\text{max}}\}\ -\min\{s_f(t),\ -2b\ v_l(t)\})/\left(v_f(t)\ +v_l(t)\right).
    \end{split}
\end{equation*}

\noindent In the following expression exploration, \textit{min} is limited to only being sampled once and should be the first token sampled. We configure a size of 300 to be the batch size of exploration. The reward penalty to encourage recommended variable interactions will be turned off after 10 epochs of exploration, i.e., \textit{EPOCH}$=10$ in Eq.~\eqref{equ:reward}, to prevent too much running time while gaining initial guidance. The classic GP method is set to be allowed to assist expression exploration. In the following, we use DSR based on the variable combinations recommended by the VIS method with the proposed complexity penalty reward strategy. All of the cases in the remainder of this subsection are tested using a noise-free dataset. We compare the performances of the pure DSR framework \cite{petersen2019deep}, the DSR framework with GP assistance \cite{larma2021improving}, and our proposed VIS-enhanced DSR framework with GP assistance. We use DSR-only, DSR-GP, and VIS-DSR-GP in the following to refer to the three methods, respectively. The DSR-GP method is configured with the same constraints of limiting \textit{min} to only being sampled once and first. Additionally, DSR-GP is configured with a constraint of limiting the length of explored expressions to a range of 10 to 40. 

\subsubsection{Expression Exploration with Penalty}
We first compare explorations using different penalty parameters of the complexity penalty strategy, i.e., $\beta$ in Eq.~\eqref{equ:reward}. In the experiment, $\beta$s are tested with values from 0.00 to 1.0, with a 0.05 increment from 0.00 to 0.20 and then with a 0.20 increment from 0.20 to 1.0. All the tests are based on the recommended variable combinations in Scenario \#1 shown in Table~\ref{tab:interactions}. Note that $\beta=0$ indicates $R=R_c$, i.e., there is no penalty if DSR explores expressions with variable combinations not recommended by the VIS.

\begin{figure}[H]
    \centering
    \begin{subfigure}{.5\textwidth}
      \centering
      \includegraphics[width=\linewidth]{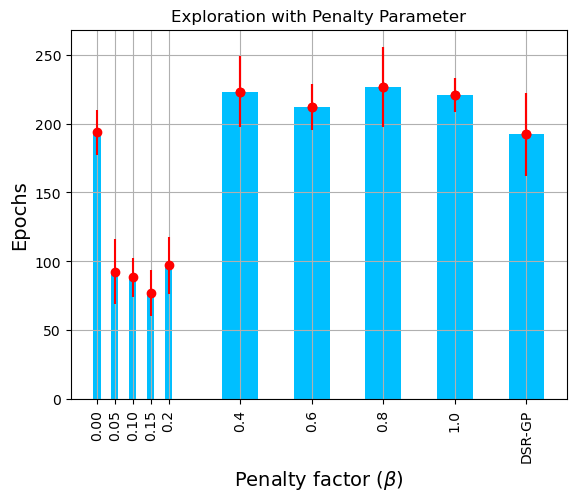}
      \caption{Reward Comparisons with Standard Errors}
      \label{fig:dsr-compare-beta-r}
    \end{subfigure}%
    \begin{subfigure}{.5\textwidth}
      \centering
      \includegraphics[width=\linewidth]{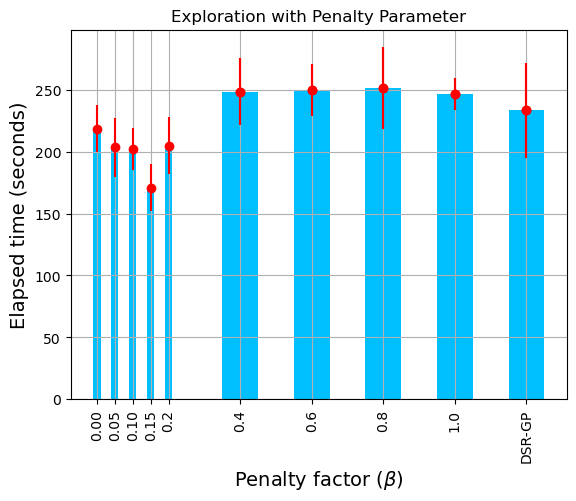}
      \caption{Elapsed Time Comparisons with Standard Errors}
      \label{fig:dsr-compare-beta-epoch}
    \end{subfigure} \\
    \begin{subfigure}{.5\textwidth}
      \centering
      \includegraphics[width=\linewidth]{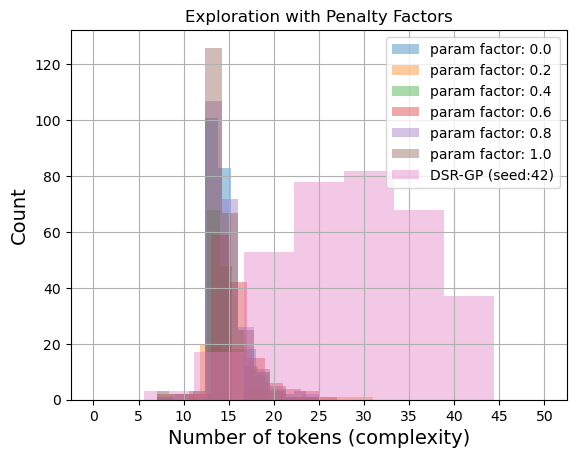}
      \caption{Density of Complexities Explored}
      \label{fig:dsr-compare-beta-comp}
    \end{subfigure}%
    \begin{subfigure}{.5\textwidth}
      \centering
      \includegraphics[width=\linewidth]{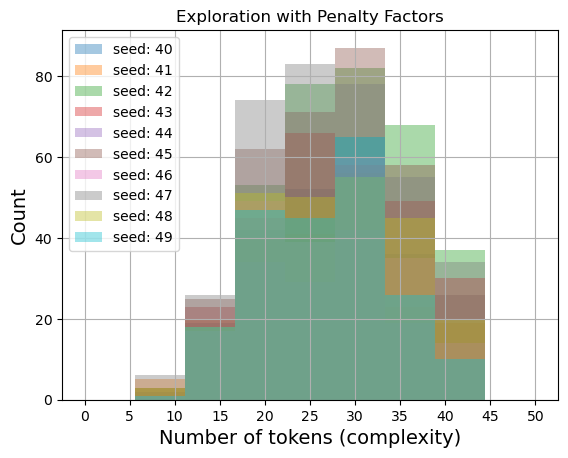}
      \caption{Density of Complexities Explored (DSR-GP)}
      \label{fig:dsr-compare-beta-seed}
    \end{subfigure}
    \caption{Exploration by Penalty Parameters}
    \label{fig:dsr-compare-beta}
\end{figure}

We compare exploration epochs until converging using the VIS-DSR-GP and DSR-GP methods with different $\beta$'s in Fig.~\ref{fig:dsr-compare-beta}, where each result is averaged over 10 random seeds with the standard errors shown by the error bars in red. All have discovered the target expression as Eq.~\eqref{equ:modified-krauss-simplified}. DSR-only is excluded from the comparison, as it does not recover the target expression.  Fig.~\ref{fig:dsr-compare-beta-r} and Fig.~\ref{fig:dsr-compare-beta-epoch} compare the numbers of learning epochs and running times over different $\beta$s, where each value is also the average of the results from running the test over 10 random seeds. The comparisons show that 0.15 is a sweet spot of $\beta$ (VIS-DSR-GP), as it takes the fewest number of epochs and the shortest running time to discover the target expression. Even adding the running time of VIS (60.6 seconds from Section~\ref{subsec:vis-result}), VIS-DSR-GP with $\beta=0.15$ is still faster than using DSR-GP, which takes more than 150 epochs and over 200 seconds to discover the target expression. Choosing $\beta=0.00$ takes more epochs to learn the dynamics, but the running time is relatively short. It is because in the test of $\beta=0.00$, the exploration does not check each expression term, and the reward takes only accuracy and parsimony into account, i.e., $R=R_c$ in Eq.~\eqref{equ:reward}. Using DSR-GP is found to tend to look for better expressions through higher complexity, as shown in Fig.~\ref{fig:dsr-compare-beta-comp} that the high range of expression complexities (20 to 40 given the desired range from 10 to 40) dominates the exploration compared to the range of low complexities (10 to 20). As shown in Fig.~\ref{fig:dsr-compare-beta-seed}, running multiple tests with the DSR-GP method, we found that this method tends to try exploring better expressions through a high level of expression complexity. Our proposed method, on the other hand, encourages the exploration of parsimonious expressions, as presented in Fig.~\ref{fig:dsr-compare-beta-comp}, where the expression complexities explored are mostly in a range of 10 to 20.

\subsubsection{Expression Exploration with Recommended Variable Combinations}
\begin{figure}[H]
    \centering
    \begin{subfigure}{.5\textwidth}
      \centering
      \includegraphics[width=\linewidth]{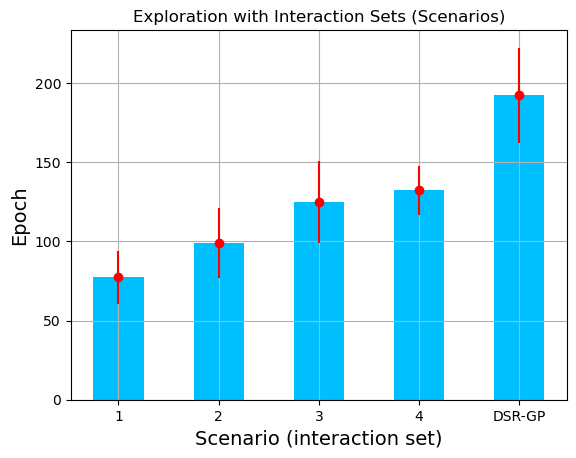}
    \caption{Exploration Epochs with Interactions}
    \label{fig:dsr-compare-epoch}
    \end{subfigure}%
    \begin{subfigure}{.5\textwidth}
      \centering
      \includegraphics[width=\linewidth]{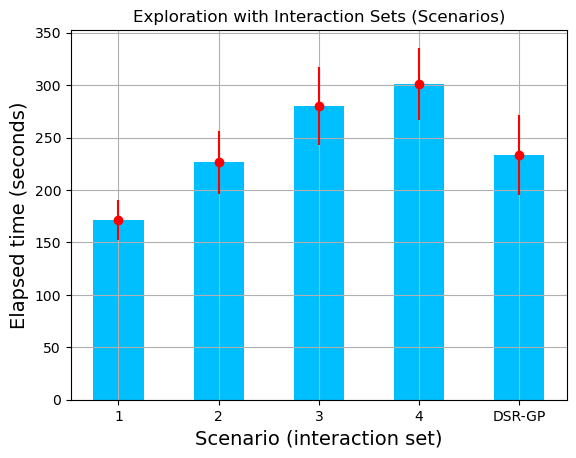}
    \caption{Run Time with Interactions}
    \label{fig:dsr-compare-elapsed-time}
    \end{subfigure}
    \caption{Exploration Performance with Variable Interactions}
    \label{fig:dsr-compare}
\end{figure}

The above experiments indicate that using $\beta=$0.15 performs the best in terms of the number of exploration epochs and running times. We adopt $\beta=0.15$ and apply the same constraints to test explorations with different recommended variable combinations listed in Table~\ref{tab:interactions}. The following tests still use the noise-free dataset. Fig.~\ref{fig:dsr-compare} presents a performance comparison using different sets of recommended variable combinations, where blue bars show the average of running 10 random seeds, and error bars in red demonstrate the standard errors. The target expression is discovered by all the test cases. The results show that a good set of variable combinations can help reduce the number of exploration epochs and further improve the running time. With that said, providing too many conditions to regulate expression terms may result in long running times, as the elapsed times of Scenarios \#2 - \#4 shown in Fig.~\ref{fig:dsr-compare-elapsed-time}. In particular, Scenarios \#3 and \#4 take a longer time to find the target expression than DSR-GP.

\subsection{Noise Impact Investigation}
\label{sec:noise-impact}

The above experiments are based on perfectly simulated data described by only one governing equation. Here we test how data noises may impact the performance of the proposed method. The following tests in this subsection are given the set of variable combinations recommended in Scenario \#1 (see Table~\ref{tab:interactions}) with $\beta=0.15$ and the same constraint configuration, and a recommended expression complexity ranging from 10 to 40.

\begin{figure}[H]
    \centering
    \begin{subfigure}{.5\textwidth}
      \centering
      \includegraphics[width=\linewidth]{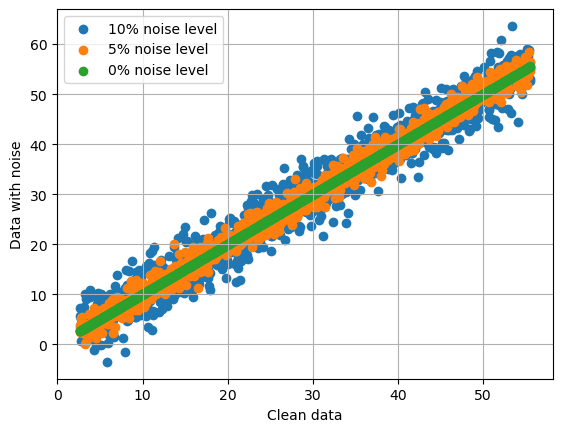}
    \caption{An Example of Data Noise}
    \label{fig:noise-example}
    \end{subfigure}%
    \begin{subfigure}{.5\textwidth}
      \centering
      \includegraphics[width=\linewidth]{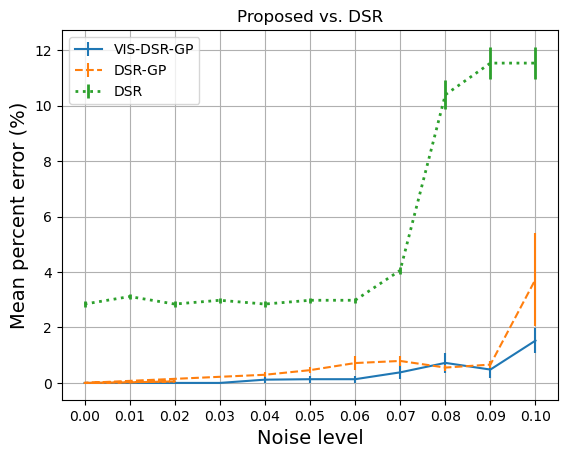}
    \caption{Epoch Exploration with Standard Errors by \\Noise Levels}
    \label{fig:dsr-compare-noise}
    \end{subfigure}
    \caption{Data Noise Example and Exploration Performance by Interactions}
\end{figure}

In the investigation, the dataset is contaminated 100\% by different noise levels, which increase from 0 to 10\% with a 1\% increment of the standard deviation of the clean data. Fig.~\ref{fig:noise-example} shows an example of the clean data, data with 5\%-level noise, and data with 10\%-level noise. The VIS algorithm yields the same recommended variable combination sets as listed in Table~\ref{tab:interactions} at all noise levels. Thus, the following focuses on comparing the performances of the DSR process under different noise levels. The performances in terms of mean percent errors (MPE) are presented in Fig.~\ref{fig:dsr-compare-noise}, where we compare the MPEs between using DSR-only, DSR-GP, and VIS-DSR-GP. 

\begin{table}[H]
\tiny
\centering
\caption{Best Recovered Expressions with Noise Levels}
\begin{tabular}{c|c|c|c}

Ground truth & \multicolumn{3}{c}{$\min \left(v_f (t)+2.6, v_l (t)+\frac{9.0 \big( s_f(t)-s_{\text{des}} (t) \big)}{ {v_f (t)+v_l (t)}+9.0} \right)$} \\ \hline
Noise level & VIS-DSR-GP    & DSR-GP                      & DSR \\ \hline\hline
0\%         & $v_l (t)+\frac{9.0 \big( s_f(t)-s_{\text{des}} (t) \big)}{ {v_f (t)+v_l (t)}+9.0} $ & $v_l (t)+\frac{9.0 \big( s_f(t)-s_{\text{des}} (t) \big)}{ {v_f (t)+v_l (t)}+9.0} $ & $v_l (t)+\frac{9.0 }{ {v_f (t)+v_l (t)}+9.0}$    \\ \hline
1\%         & $v_l (t)+\frac{9.0 \big( s_f(t)-s_{\text{des}} (t) \big)}{ {v_f (t)+v_l (t)}+9.0} $ & $v_l (t)+\frac{9.0 \big( s_f(t)-s_{\text{des}} (t) \big)}{ {v_f (t)+v_l (t)}+9.0} $ &  $v_l (t)+\frac{1}{9.0\left(s_f(t)-v_l(t)\right)}$   \\ \hline
2\%         & $v_l (t)+\frac{9.0 \big( s_f(t)-s_{\text{des}} (t) \big)}{ {v_f (t)+v_l (t)}+9.0} $ & $v_l (t)+\frac{9.0 \big( s_f(t)-s_{\text{des}} (t) \big)}{ {v_f (t)+v_l (t)}+9.0} $ &  $v_l (t)+\frac{\left(2s_f(t) + v_l(t)\right)}{9.0} $   \\ \hline
3\%         & $v_l (t)+\frac{9.0 \big( s_f(t)-s_{\text{des}} (t) \big)}{ {v_f (t)+v_l (t)}+9.0} $ & $v_l (t)+\frac{9.0 \big( s_f(t)-s_{\text{des}} (t) \big)}{ {v_f (t)+v_l (t)}+9.0} $ &  $v_l(t)+\frac{v_f(t)}{ v_f (t)+2.6} - 2.6 $   \\ \hline
4\%         & $v_l (t)+\frac{9.0 \big( s_f(t)-s_{\text{des}} (t) \big)}{ {v_f (t)+v_l (t)}+9.0} $ & $v_l (t)+\frac{9.0 \big( s_f(t)-s_{\text{des}} (t) \big)}{ {v_f (t)+v_l (t)}+9.0} $ &  $v_l(t) + \frac{s_{\text{des}}-9.0}{v_f(t)+9.0}$   \\ \hline
5\%         & $v_l (t)+\frac{9.0 \big( s_f(t)-s_{\text{des}} (t) \big)}{ {v_f (t)+v_l (t)}+9.0} $ & $v_l (t)+\frac{9.0 \big( s_f(t)-s_{\text{des}} (t) \big)}{ {v_f (t)+v_l (t)}+9.0} $ & $v_l(t)+\frac{s_{\text{des}}}{9.0}+0.3$    \\ \hline
6\%         & $v_l (t)+\frac{9.0 \big( s_f(t)-s_{\text{des}} (t) \big)}{ {v_f (t)+v_l (t)}+9.0} $ & $v_l (t)+\frac{9.0 \big( s_f(t)-s_{\text{des}} (t) \big)}{ {v_f (t)+v_l (t)}+9.0} $ & $v_l(t)+\frac{s_{\text{des}}}{9.0}+0.3$    \\ \hline
7\%         & $v_l (t)+\frac{9.0 \big( s_f(t)-s_{\text{des}} (t) \big)}{ {v_f (t)+v_l (t)}+9.0} $ & $v_l (t)+\frac{9.0 \big( s_f(t)-s_{\text{des}} (t) \big)}{ {v_f (t)+v_l (t)}+9.0} $ &  $v_l(t)+\frac{s_{\text{des}}}{9.0}-1.0$   \\ \hline
8\%         & $v_l (t)+\frac{9.0 \big( s_f(t)-s_{\text{des}} (t) \big)}{ {v_f (t)+v_l (t)}+9.0} $ & $v_l (t)+\frac{9.0 \big( s_f(t)-s_{\text{des}} (t) \big)}{ {v_f (t)+v_l (t)}+6.4} $ & $v_l(t)-2.6$   \\ \hline
9\%         & $v_l (t)+\frac{9.0 \big( s_f(t)-s_{\text{des}} (t) \big)}{ {v_f (t)+v_l (t)}+9.0} $ & $v_l(t) \left(1 -\frac{s_f(t)+9.0}{s_f(t)\bigg(s_f(t)+9.0\bigg)+v_f(t)+v_l(t)+9.0}\right) $ & $v_l(t)-2.6$   \\ \hline
10\%        & $v_l (t)+\frac{9.0 \big( s_f(t)-s_{\text{des}} (t) \big)}{ {v_f (t)+v_l (t)}+9.0} $ & $v_l (t) \left(1 - \frac{9.0}{v_f(t) + v_l(t) + 9.0 s_f(t)}\right) $ &   $v_l(t)-2.6$  \\ 
\end{tabular}
\label{tab:expr-uncover-all-methods}
\end{table}

This figure shows MPEs (with respect to the ground truth) over different noise levels, where each result is the average of running 10 random seeds with error bars (vertical lines) indicating standard errors. The results demonstrate that DSR-only fails to discover a suitable expression that can adequately capture the behaviors performed in the data for all cases, even on clean data (0\% noise level). The average MPEs are at around 3\% given the noise levels below 7\%. DSR-GP can reconstruct the target expression for all random seeds on clean data. The averaged MPEs drop below 2\% for noise levels from 0\% to 9\%, while MPEs range 2-5\% when the dataset is imposed 10\% of noise. The results clearly show that GP helps reduce the impact of data noise. On the other hand, with prior knowledge of potential variable combinations, VIS-DSR-GP can uncover the target expression for all tests when noise levels are at or below 3\%, and produce expressions with the highest MPEs below 2\% (at 10\% of noise level). The best expressions (out of the 10 random seeds) recovered using these methods are listed in Table~\ref{tab:expr-uncover-all-methods}, where we only document the component of the second term in the \textit{min} function, as all the methods can recover the first term ($v_f(t)+2.6$). Here the expression is chosen as \textit{the best} if it has the highest accuracy and/or the most similar expression structure as the target expression. The results show that DSR-only is not able to find the correct expression at all noise levels. It falls into a local optimum, where the expression structure of each term is limited. With the help of GP, DSR-GP can recover the target expression at or below 6\%. On the other hand, VIS-DSR-GP can recover the target expression for all tests at least once at all noise levels. The results also show that learning the exact structure of the expression (a division of two terms in this case) and the correct values of the constant (e.g., 9.0) are challenging for DSR and DSR-GP.

\subsection{Testing Performance on Other Car-following Models}
\label{sec:additional-study}
In this subsection, we test the expression reconstruction capability of the proposed VIS-enhanced DSR framework (VIS-DSR-GP) on other car-following models. We choose the GM model \citep{chandler1958traffic} and the Gazis-Herman-Rothery (GHR) model \citep{gazis1961nonlinear} as the target expressions in the following experiments. The two expressions are respectively listed in Eq.~\eqref{equ:chandler_dis} and Eq.~\eqref{equ:ghr}. Here we simplify the sensitivity term $c$ in the GM model as a constant.

\begin{equation}
    v_{f} (t+\delta t) = v_{f} (t) + c[v_l (t) - v_f (t)].
    \label{equ:chandler_dis}
\end{equation}


\begin{equation}
    v_{f} (t+\delta t) = v_{f} (t) + k_1{v}_{f}^{k_2}(t)
    \frac{v_l(t-\delta t) - v_f(t-\delta t)}{s^{k_3}_f(t-\delta t)}.    
    \label{equ:ghr}
\end{equation}

Each car-following dataset is generated through the SUMO numerical simulation with the corresponding car-following model sharing the same vehicle parameters, including $\delta t$, $a_{\text{max}}$, $v_{\text{max}}$, $b$, and $t_{\text{react}}$, as the Krauss model mentioned in Section \ref{subsec:dataset}. For constants in the tested car-following models, we use the recommended $c=0.368$ \citep{chandler1958traffic} for the GM model, and $k_1=1.2$, $k_2=1.0$, $k_3=1.1$ for the GHR model \citep{al2009examining, may1967non}. While collecting the GHR car-following data, we simplify the velocity difference as a term, i.e., $\Delta v(t-\delta t)= v_l (t-\delta t) - v_f (t-\delta t)$. The simplified car-following models with pre-determined parameters are as Eq.~\eqref{equ:chandler_dis-sim} and Eq.~\eqref{equ:ghr-sim} for GM and GHR, respectively.

\begin{equation}
    v_{f} (t+1) = v_{f} (t) + 0.368 \left[v_l (t) - v_f (t)\right].
    \label{equ:chandler_dis-sim}
\end{equation}

\begin{equation}
    v_{f} (t+1) = v_{f} (t) + 1.2{v}_{f}(t)
    \frac{\Delta v(t-1)}{s^{1.1}_f(t-1)}.
    \label{equ:ghr-sim}
\end{equation}

\begin{figure}[H]
    \centering
    \begin{subfigure}{.5\textwidth}
      \centering
      \includegraphics[width=\linewidth]{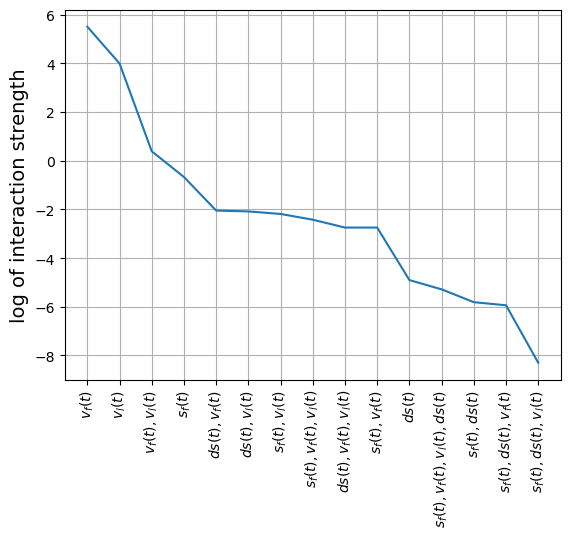}
    \caption{GM Model}
    \label{fig:gm-is}
    \end{subfigure}%
    \begin{subfigure}{.5\textwidth}
      \centering
      \includegraphics[width=\linewidth]{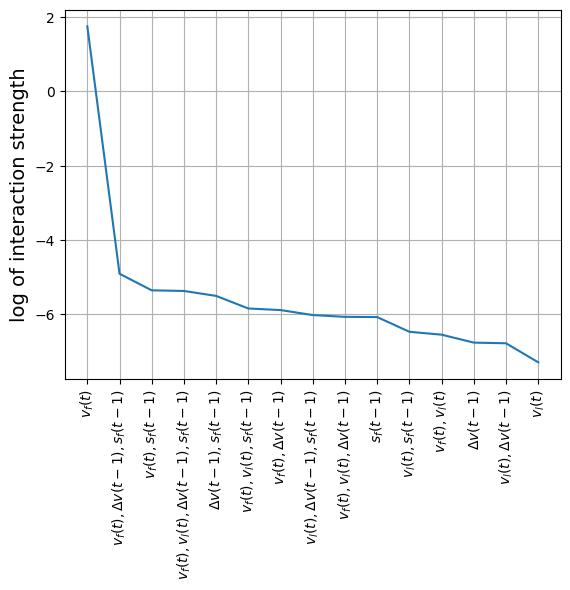}
    \caption{GHR Model}
    \label{fig:ghr-is}
    \end{subfigure}
    \caption{Interaction Strengths}
    \label{fig:ghr-gm-is}
\end{figure}

The collected GM dataset shares the same variables as the Krauss model dataset, i.e., features in the GM dataset include $v_f(t)$, $v_l(t)$, $s_f(t)$, and $ds(t)$. The collected GHR dataset, on the other hand, comprises variables $v_f(t)$, $v_l(t)$, $s_f(t-1)$, and $\Delta v(t-1)$, as it has more specific features. Feeding both datasets respectively in the VIS algorithm obtains interaction strengths as shown in Fig.~\ref{fig:ghr-gm-is}. We apply the elbow method to determine recommended combinations and document them in Table~\ref{tab:interactions-other}. 

Now, we use the VIS-enhanced DSR framework to reconstruct the target expressions. After being simplified, the GM and GHR expressions have one and two constants, respectively. We reconstruct the two models by giving the framework the following mathematical token pool: \{$+$, $-$, $\times$, $\div$, $\min$, power, constant\},  parameters \{$a_{\text{max}}$, $v_{\text{max}}$, and $b$\}, and the variables as shown in Table \ref{tab:interactions-other}. The hard constraints include: i) \textit{min} and \textit{constants} cannot occur more than twice, and ii) \textit{min} should be the first sampled token. Soft regulations, such as the reward penalty factor and recommended complexity, are configured the same as the values in the previous subsections.

\begin{table}[H]
\scriptsize
\centering
\caption{Variable Interaction Strengths}
\begin{tabular}{c|c|c}
             & GM                                                                             & GHR \\ \hline
Interactions & \begin{tabular}[c]{@{}l@{}}$v_f(t)$\\ $v_l(t)$\\ $(v_f(t), v_l(t))$\end{tabular} & \begin{tabular}[c]{@{}l@{}}$v_f(t)$\\ $(v_f(t),\Delta v (t-1),s_f(t-1))$\end{tabular}  
\end{tabular}
\label{tab:interactions-other}
\end{table}

By imposing noise from level 0\% to 10\%, we document the best-reconstructed expressions in Table~\ref{tab:expr}, to which we document only the second term in each expression since the first term can be correctly learned by the method for all noise levels. From the table, VIS-DSR-GP is able to reconstruct the GM expression with the given token pool and constraints for the zero-noise dataset. Although the decimal values of the explored constants between noise levels 1\% to 7\% do not exactly match the target value, they are quite close (less than 0.5\%). However, the constant starts to differ significantly from the target value when the noise level exceeds 8\%. For the GHR, on the other hand, the best expression the framework can explore is $v_{f} (t+1) = v_{f} (t) + 1.01{v}_{f}(t) \frac{v_l(t-1) - v_f(t-1)}{s_f(t-1)-0.16}$. Although the resulting expression has the same structure as the GHR: $v_f(t)$, plus the product of a constant and a fraction term, the constant and the power term (also a constant) in the denominator is the most challenging to learn and discover. This becomes even more challenging when the data has a higher level of noise. Starting at 5\% noise level, an additional term appears beside the primary expression structure. When the noise level exceeds 9\%, the structure of the explored expression begins to noticeably differ from the target expression of GHR. The results here further confirms that learning the exact structure of the model and the constants (especially those with decimal values) are challening even for the proposed VIS-DSR-GP method.

\begin{table}[H]
\scriptsize
\centering
\caption{Best Recovered Expressions Using VIS-DSR-GP}
\begin{tabular}{c|c|c}

            & GM                      & GHR \\ \hline
Noise level & $0.368\left(v_l(t) - v_f(t)\right)$ &  $1.20{v}_{f}(t) \frac{\Delta v(t-1)}{s^{1.1}_f(t-1)}$   \\ \hline\hline
0\%         & $0.368\left(v_l(t) - v_f(t)\right)$ & $1.01{v}_{f}(t) \frac{\Delta v(t-1)}{s_f(t-1)-0.16}$    \\ \hline
1\%         & $0.367\left(v_l(t) - v_f(t)\right)$ &  $1.01{v}_{f}(t) \frac{\Delta v(t-1)}{s_f(t-1)-0.16}$   \\ \hline
2\%         & $0.366\left(v_l(t) - v_f(t)\right)$ &  $1.01{v}_{f}(t) \frac{\Delta v(t-1)}{s_f(t-1)-0.16}$   \\ \hline
3\%         & $0.366\left(v_l(t) - v_f(t)\right)$ &  $1.01{v}_{f}(t) \frac{\Delta v(t-1)}{s_f(t-1)-0.16}$   \\ \hline
4\%         & $0.366\left(v_l(t) - v_f(t)\right)$ &  $1.00{v}_{f}(t) \frac{\Delta v(t-1)}{s_f(t-1)-0.16}$   \\ \hline
5\%         & $0.366\left(v_l(t) - v_f(t)\right)$ & $1.00{v}_{f}(t) \frac{\Delta v(t-1)}{s_f(t-1)-0.17}+\frac{0.21}{\Delta v(t-1)}$    \\ \hline
6\%         & $0.366\left(v_l(t) - v_f(t)\right)$ & $1.00 \frac{\Delta v(t-1)}{s_f(t-1)-0.17}-0.01v_f(t)$    \\ \hline
7\%         & $0.366\left(v_l(t) - v_f(t)\right)$ &  $1.00{v}_{f}(t) \frac{\Delta v(t-1)}{s_f(t-1)-0.17}+\frac{0.02}{2\Delta v(t-1)}$   \\ \hline
8\%         & $0.385\left(v_l(t) - v_f(t)\right)$ & $1.00 v_f(t)\frac{\Delta v(t-1)}{s_f(t-1)-0.17}-0.78$    \\ \hline
9\%         & $0.385\left(v_l(t) - v_f(t)\right)$ &  $v_f(t)\Delta v(t-1)(-0.06+\frac{1.27}{s_f(t)})$   \\ \hline
10\%        & $0.385\left(v_l(t) - v_f(t)\right)$ &   $\frac{\left[v_f(t)\left(\Delta v(t-1)\right)\right]^2}{v_f(t)\Delta v(t-1)\left(s_f(t)-0.06\right)-0.57}$  \\ 
\end{tabular}
\label{tab:expr}
\end{table}

\section{Discussions}
\label{sec:discussion}
The results of the above numerical experiments illustrate that under relatively simple scenarios - Krauss or GM model, single vehicle class, noise-free or low-noise level (less than 8\% as shown in Table \ref{tab:expr-uncover-all-methods} and Table \ref{tab:interactions-other}) - the proposed VIS-enhanced DSR method can discover the target car-following models exactly or almost exactly from data directly with minimum human involvement. We think this is quite impressive considering the various structures/terms, math operations, variables, and especially constants in the expressions of the car-following models. This is especially so when compared with traditional methods for developing car-following models, for which extensive, time-consuming, and iterative processes of data analysis, visualization, hypothesis, validation, and testing are essential steps. The results, however, also demonstrate that the proposed framework faces significant challenges in exploring expressions correctly for dynamics with complex expression structures, constants, and data noises. Target expressions with intricate structures, such as the power term in the GHR model, can make it more difficult for the proposed method to find the correct expression. When there are many constants to be fitted, finding the exact expressions can also be challenging. Another challenge is data noises, which can have a significant impact on the exploration process, as they can obscure the desired predictions and thus make it difficult to obtain the target expression terms correctly. Furthermore, real-world traffic streams often contain multiple vehicle classes, such as cars, buses, and trucks, which may have different key model parameters (maximum speed, acceleration, etc.) even if they follow the same car-following dynamics. Our initial testing shows that this can bring great challenge to the proposed learning method, due to the large number of constants and the multimodal nature of the problem. As a result, there is much room for improvements in VIS-enhanced DSR to address these challenges.

More advanced deep learning techniques, such as large language model (LLM) based methods, need to be explored to address these challenges; see Section \ref{sec:conclusion} for more discussions on this aspect. On the other hand, clearly defining objectives and developing measures that can properly capture such objectives are also crucial. With the goal of learning the governing equations of dynamical systems, such as the car-following dynamics in this paper, the learned mathematical expressions should accurately capture the observed data, which should also have proper physical meanings and can be generalized to different sites and traffic conditions. Learning accuracy is not the only objective, which is starkly different from other DL-based applications. This prompts the definition of four key measures that help evaluate the quality of the derived expressions: i) accuracy, ii) interpretability, iii) parsimony, and iv) generalizability, where interpretability, parsimony (compactness), and generalizability are also discussed in \citep{tenachi2023deep}.

\textbf{Accuracy} measures the “goodness of fit” of the learned expressions with respect to the observed data, which is the basic measure to evaluate the quality of the learned expression. A good expression should provide reasonable accuracy in fitting the observed data. Metrics such as MSE, NMSR, or MPE have been widely used to measure accuracy; for example, DSR uses NMSE as one measure for the performance of the resulting expression. In addition, other mathematical properties of the model, such as stability, are also studied and used as metrics to evaluate the performance of the models. 

\textbf{Interpretability} means that a learned expression should have human-understandable justifications about the “physical meaning” of the model. It is important as it implies “trust” – people often have more confidence in an interpretable model that can then be more easily adopted and used \citep{murdoch2019definitions, arrieta2020explainable, madiega2021artificial}.  Interpretability is an important objective for learning governing equations, as the main goal is to add new knowledge to our understanding of the physical systems by developing such models. This can be particularly essential (and challenging) when we learn the potentially new dynamics for mixed traffic flow of CAVs and HDVs, as discussed in the Conclusion Section below. As we expect that explicit analytical expressions are produced in this research, we focus on two aspects of interpretability: i) the variables in the model represent clear physical phenomena or characteristics of the traffic system; and ii) the relations of the variables described in a model have clear, concise (see parsimony below) explanations. In this paper, we applied VIS to find the key (strong) interactions among variables, with the help of constraining the model complexity, which helps produce expressions that have reasonable physical meanings. As interpretability and the above two aspects are more subjective, it is imperative to develop specific measures for interpretability that can be properly integrated into the DSR learning process (e.g., the design of the reward function). 

\textbf{Parsimony} is “a good principle to explain the phenomena by the simplest hypotheses possible” \citep{GauchJr_2002,lycan1975occam} or the simplest possible solution to a problem, which has been widely used as a guiding principle for daily life and scientific investigations \citep{lycan1975occam,bentley1993engineering}. In our specific context, parsimony means that the learned expressions are concise and straightforward, given all others, e.g., the metrics for other objectives being equal or very similar. In this paper, we use the complexity of the expression, i.e., the number of variables and operations included in the expression, as the measure to define and evaluate its parsimony. However, parsimony is also a subjective measure requiring substantial human judgment of the learned expression. Specific procedures/measures need to be developed to help evaluate the parsimony of an expression so that it is not too complex or over-parsimonious (i.e. too simple), and also help reach a proper balance with other objectives. 

\textbf{Generalizability} means that the learned governing equations can be applied to data collected from broader traffic systems at various sites with different traffic mixes and conditions, which indicates that the learned expression can capture the underlying dynamics of the traffic system rather than only fitting to a specific dataset. Generalizability can be measured by the model performance (e.g., accuracy) across different datasets and different sites. An expression that can describe all scenarios better is considered a more generalizable model than those that only work well for certain datasets but not on others. This may be measured by the metrics such as their averages and variances of other three objectives discussed above across different datasets and test sites. Furthermore, expressions with no or minimum calibration but can well describe the dynamics of different datasets and/or sites also have better generalizability properties.

\section{Conclusion}
\label{sec:conclusion}

In this study, we proposed an expression exploration framework based on the DSR approach while encouraging parsimonious mathematical expressions that include potential variable combinations detected through a VIS algorithm. The variable combinations are used to guide DSR in focusing on expressions that involve the identified combinations, which can also increase the interpretability of the explored mathematical expressions. For this, we designed two penalty terms for the reward function of training RNNs in DSR: (i) a complexity penalty to encourage the exploration to crawl toward the desired range of expression complexity, and (ii) a variable interaction penalty to encourage the exploration of expressions consisting of the recommended variable combinations. The following results were discovered. First, the recommended variable combinations were found useful for directing the exploration to identify suitable mathematical expressions. Secondly, the proposed complexity penalty strategy was shown to be better for regulating the complexity range searched than the traditional complexity regulation method in DSR. Thirdly, the exploration based on the recommended variable combinations with the complexity penalty strategy helped identify the governing equations more efficiently than the traditional DSR. From the data noise investigation experiments, we recommended turning on the exploration assistance of GP, as it enables a wider search space for exploration. 

The methods and results in this paper present the first step of discovering governing equations of car-following dynamics using deep learning. Limitations exist which motivate future research. We need to improve the current learning framework to address the limitations summarized in Section \ref{sec:discussion}, and apply the improved framework to more diversified, real trajectory data. In particular, the proposed framework can be applied to multimodal traffic systems, e.g., to capture the car-following dynamics among CAVs and between HDVs and CAVs, and to help develop interpretable dynamical models for the interactions between vehicles and vulnerable road users. This may be done by integrating multiple dynamic exploration frameworks as a coordinated learning system in which each exploration framework learns an interaction type in a coordinated way from the collected multimodal data. 

There are two main directions that can help improve the current learning framework and address the challenges. First, as discussed in the previous section, accurately fitting the observed data is NOT the only objective for deriving proper governing equations. One has to balance with other objectives, including interpretability, parsimony, and generalizability. Proper metrics for each of these objectives need to be defined and integrated into the learning process, e.g., to design a comprehensive reward function that includes metrics for measuring parsimony, interpretability, generalization, and accuracy. An exploration incorporated with a reward that takes these features into account may also help find appropriate expressions from data with noise and complicated underlying dynamics. However, developing such a reward function is challenging, as there may be conflicts between accuracy and other objectives. Despite the challenges, this is an important research topic that has the potential to advance the capability of using deep learning models to "automatically" discover the underlying expressions from collected data. Second, given the advance of deep learning especially recent LLMs, one should leverage the ability of interaction and memory of LLMs to enhance search efficiency and expression quality in discovering the governing equations. Research has started in this direction \cite{kamienny2022end}, which needs to be further improved and tailored to car-following and other engineering applications.

\bibliographystyle{unsrtnat}
\bibliography{references}  






\end{document}